\pdfoutput=1
\UseRawInputEncoding
\documentclass[runningheads]{llncs} \usepackage[T1]{fontenc}
\usepackage{graphicx}
\usepackage{soul,color}
\usepackage{verbatim}
\usepackage{url}

\begin{document}
\title{Analyzing Deep Learning Based Brain Tumor Segmentation with Missing MRI Modalities}
\titlerunning{Brain Tumor Segmentation with Missing MRI}
%
\author{Benteng Ma\inst{1} \and
Yushi Wang\inst{1} \and
Shen Wang\inst{2}}
\authorrunning{B. Ma et al.}
%
\institute{
Beijing-Dublin International College, University College Dublin, Dublin, Ireland\\
\email{\{benteng.ma,yushi.wang\}@ucdconnect.ie}
\\
 \and
School of Computer Science, University College Dublin, Dublin, Ireland
\email{shen.wang@ucd.ie}
}
\maketitle              
\begin{abstract}
This technical report presents a comparative analysis of existing deep learning (DL) based approaches for brain tumor segmentation with missing MRI modalities. Approaches evaluated include the Adversarial Co-training Network (ACN) \cite{wang2021acn} and a combination of mmGAN \cite{sharma2019missing} and DeepMedic \cite{kamnitsas2017efficient}. A more stable and easy-to-use version of mmGAN is also open-sourced at a GitHub repository\footnote{\url{https://github.com/MBTMBTMBT/mmGAN_Tensorflow}\label{ft1}}. Using the BraTS2018 dataset, this work demonstrates that the state-of-the-art ACN performs better especially when T1c is missing. While a simple combination of mmGAN and DeepMedic also shows strong potentials when only one MRI modality is missing. Additionally, this work initiated discussions with future research directions for brain tumor segmentation with missing MRI modalities.

\end{abstract}

\section{Introduction}

A high-quality magnetic resonance imaging (MRI) scan is of vital importance to the downstream workflows as such diagnosis and treatment plans. The segmentation of MRI images has been a well-known challenge for a long time. Recently, deep-learning-based approaches such as DeepMedic\cite{kamnitsas2017efficient} have improved such segmentation results to comparable human-level performance. However, in practice, we cannot assume the MRI images are always high-quality. This is because these images often contain defects for one or more MRI sequences (e.g., T1, T2, Fair, etc.). Existing deep-learning-based approaches such as mmGAN\cite{sharma2019missing} can well synthesize one or more missing MRI sequences to complement the reduced performance. Some recent attempts like\cite{wang2021acn} have reaffirmed the potential of using deep learning for brain tumor segmentation under various missing MRI sequences. However, there are still some open questions such as is a straightforward combination of mmGAN and DeepMedic (mmDM) good enough? Is ACN practically usable? Does a better missing MRI sequence synthesis always lead to a better brain tumor segmentation? This technical report provides our views on these questions by a comprehensive evaluation of ACN and mmDM. We also provide our Tensorflow implementation, open-sourced at \textsuperscript{\ref{ft1}}, of mmGAN to facilitate this evaluation. 

\section{Our Implementation of mmGAN}
\subsection{Introduction of mmGAN}
\begin{figure}[htb]
\includegraphics[width=\textwidth]{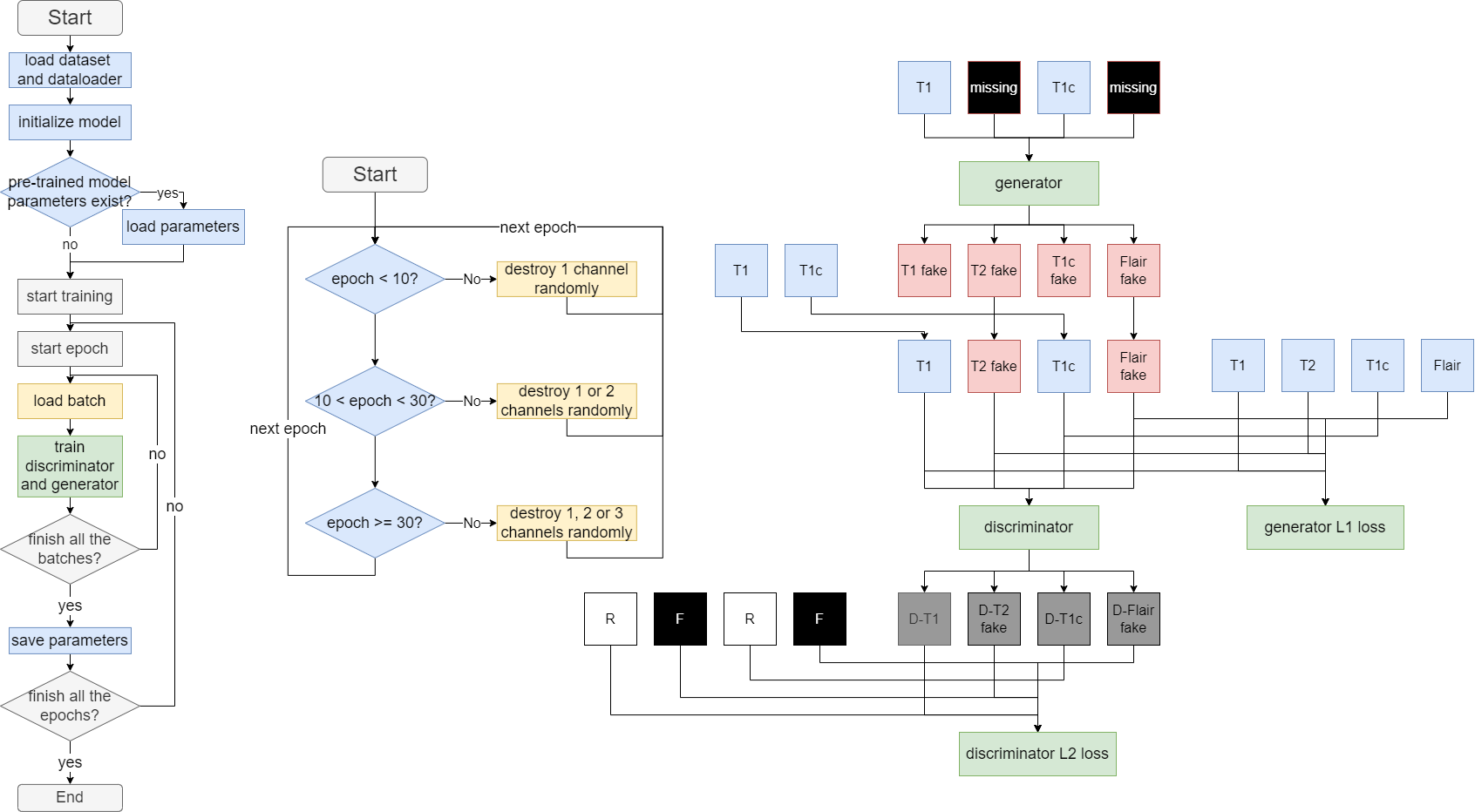}
\caption{ The work flow of training mmGAN (left), curriculum learning (centre), and implicit conditioning (right) \cite{sharma2019missing}} 
\label{fig:ori_flowchart}
\end{figure}

Multi-Modal Generative Adversarial Network (mmGAN) \cite{sharma2019missing} can generate one or more missing MRI sequences using only one generator model, with at least one sequence as input, through one forward propagation. mmGAN is based on U-Net\cite{ronneberger2015u}, which has been demonstrated to have strong capabilities in segmentation tasks with limited amount of medical images data set.
\noindent
As shown in the leftmost part of Figure \ref{fig:ori_flowchart}, the image channels are fetched and trained batch by batch during each epoch. The curriculum learning shown in the central part of Figure \ref{fig:ori_flowchart} is achieved by taking different random numbers of bad channels during each epoch, more channels will be randomly destroyed when the number of epoch increases. To achieve implicit conditioning shown in the rightmost part of Figure \ref{fig:ori_flowchart}, original images of the input channels (good channels, not the bad ones) will be kept and replace the generated ones during training, so that the discriminator will only make the decision based on the generated lost (bad) channels.

\subsection{Our improvements to the original mmGAN implementation}
Comparing with the original mmGAN implementation\footnote{\url{https://github.com/trane293/mm-gan}\label{ft2}}, our version provides more options for data preprocessing and training, as well as makes the separation of the dataset easier, and improves the execution efficiency. Users may have more flexibility using the network, and can easily use it with other downstream tasks.

\paragraph{A Tensorflow version to better cooperate with DeepMedic}
The original mmGAN implementation is based on PyTorch. We have our mmGAN implementation based on TensorFlow to give researchers another choice and also to better integrate with DeepMedic for the MRI brain tumor segmentation with missing MRI sequences.

\paragraph{Easier configuration and command-line usage}
The current implementation uses a complex configuration file for various functions. It is difficult for users to understand its contents, or modify it according to local environments. The original authors' version of the code mainly focuses on experiments, a lot of the components are used for preprocessing different datasets or doing measurements, which might cause confusion when read by other users of the code. The users may also have difficulties in using different datasets rather than BRATS18, or utilizing further the outcome of the network.
\noindent
We allow users to modify the preprocessing and training configurations easily, and make the model adjustable to different sequential datasets. All the processes are controllable through the command line, bringing convenience for batch processing. Users may first call preprocessing function to preprocess the data, and then train the model with either TensorFlow or PyTorch frameworks, after that they may use the test function to predict certain groups of images with all the missing sequence scenarios. The synthesised results can be used for downstream tasks such as tumor segmentation.

\paragraph{Easier achievement of cross validation}
Using the original implementation, it can be difficult to manage data and achieve cross-validation, datasets have to be managed manually instead of automatically. With our implementation, users can easily separate the dataset into several folds and select which folds are used for training or validation.

\paragraph{Allowing random data sequences for better performance}
In our code, we allow the slices of images of different patients to be shuffled randomly, thus the slices in each batch come from different patients, without certain order. The purpose is not only to avoid the model bias on certain sequences of continued slices but also to relieve the negative effect of continued damaged sequences that appears in the dataset. Additionally, the user may also choose to use different combinations of missing sequences within one batch, by changing the ``full\_random'' parameter. For example, when setting the parameter into ``False'', a batch with the size of 8 will have all the 8 groups of MRI images with the same channels to be set to zero; yet with the ``True'' setting, each group would have different channels to be the missing ones.

\begin{figure}[htb]
\includegraphics[width=0.9\textwidth]{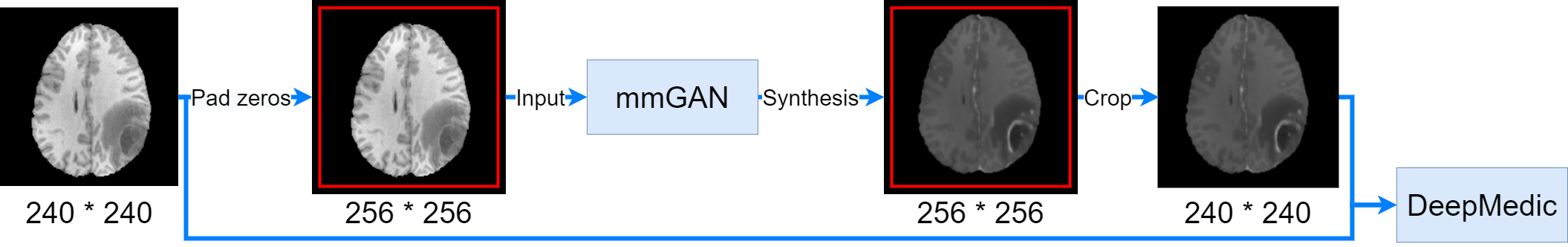}
\caption{Process of padding the inputs} 
\label{fig:zero_padding}
\end{figure}

\begin{figure}[htb]
\includegraphics[width=0.9\textwidth]{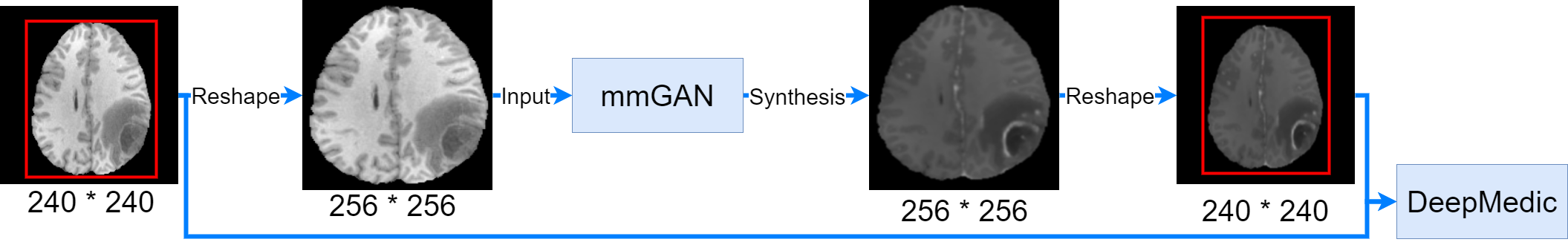}
\caption{Process of cropping and reshaping the inputs} 
\label{fig:cropping}
\end{figure}

\paragraph{Support downstream (DeepMedic) usage}
The image size of BRATS datasets is 240*240, the same as some tumor segmentation models' requirements (such as DeepMedic), yet the input size of mmGAN is 256*256. One method is to pad (as shown in Figure \ref{fig:zero_padding}) the BRATS images directly with zeros to the expected resolution, use them for synthesis, and then crop the 240*240 region out for DeepMedic; another is to crop the images with a bounding box and resize them into 256*256 to the dataset (as shown in Figure \ref{fig:cropping}), using them for training, as \cite{sharma2019missing} did, then resize it back, 
to the original resolution to suit the requirement of DeepMedic. The user may change the ``operation'' parameter between ``padding'' and ``crop''.

\paragraph{Better efficiency for preprocessing}

During preprocessing, the original implementation requires around 60 GB of memory, which is unnecessary, and beyond the memory capacity of most PCs; the generated HDF5 files are not compressed, which also occupy large space in the disk; the data loader used during training may not have good support of multi-processing in some environments, but running within only one process it takes much more time for training.
\noindent
We remake the preprocess functions, compare to the original version, our version is much more memory efficient. The preprocess program runs with multi processes. For each patient, a mean of all the voxels (within a bounding box of the main region of the brain) is computed, then the value of each voxel is divided by this mean value for standardization. Images of only one patient are opened in the memory once, so there is no special requirement of memory size; because of the parallelism, time efficiency is ensured.
\noindent
We use the TensorFlow TF-Record dataset, which requires less memory and has better support for multi-processing. The method allows data to be compressed to take up less space in the disk but is still highly efficient to be loaded. The provided data loader of TensorFlow will automatically choose the number of processes to use and adapt the speed to increase efficiency. When running preprocessing functions, the images from patients will be arranged randomly into several TF-Record files, assigned in the configuration, some can be used as test and validation datasets and the others can be used for training.

\section{Experiment Results and Analysis}
This section explains the experiment methodology, presents the experiment results along with their analysis. All experiments are using the BraTS2018 dataset \cite{menze2014multimodal} for training, validating, and testing.

\subsection{mmGAN: original v.s. ours}
\begin{figure}[htb]
\centering
\includegraphics[width=0.8\textwidth]{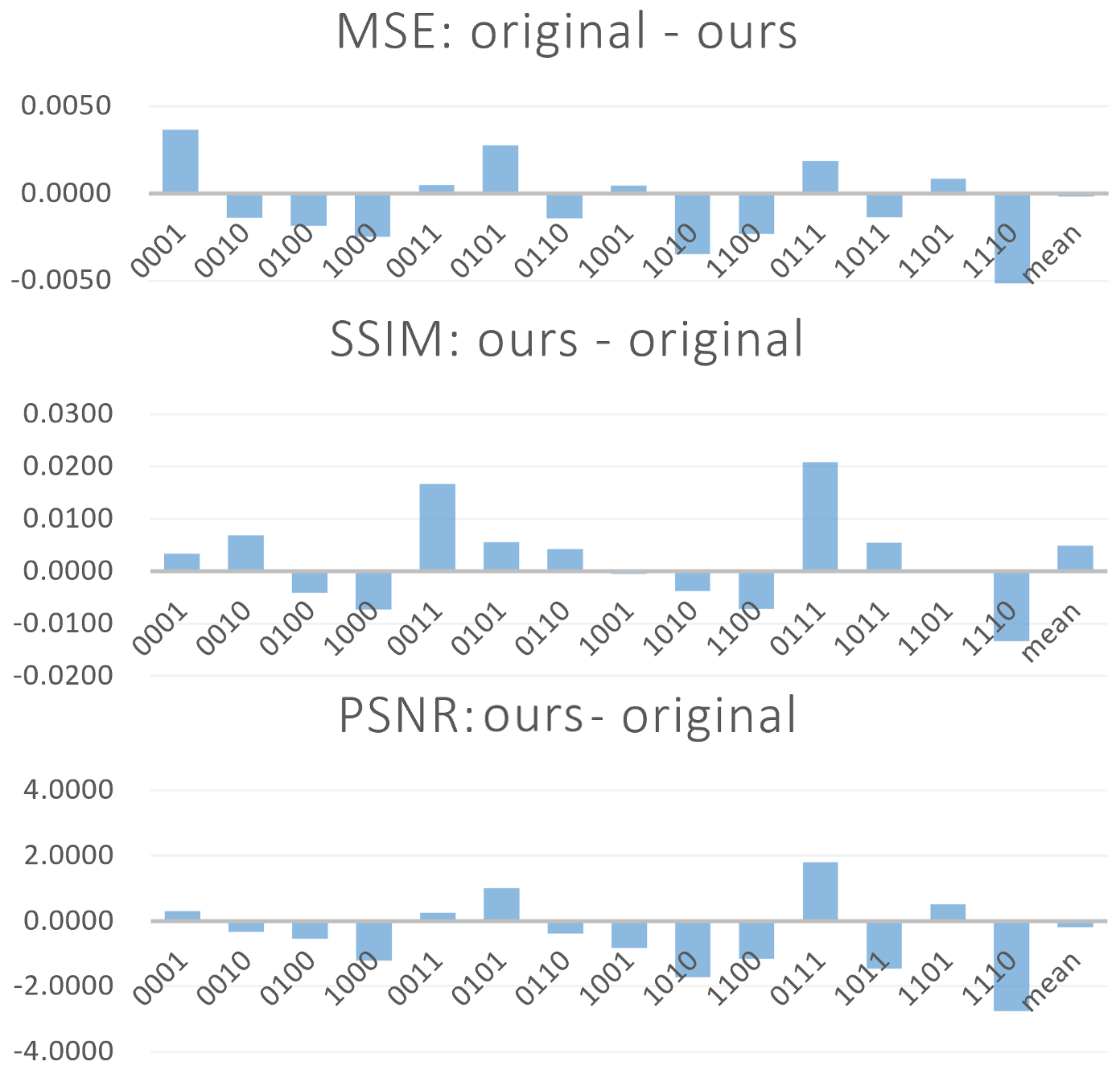}
\caption{The comparison of our implementation to the original mmGAN implementation across three different metrics MSE, SSIM, PSNR. We show the values of the difference between the original and ours: original minus ours for MSE, ours minus original for SSIM and PSNR. Higher values mean that our implementation leads to better results. Overall, the results are quite close to zero, which means that our implementation achieves similar results as the original does.} 
\label{fig:mse_ssim_psnr_diff}
\end{figure}

\begin{figure}[htb]
\includegraphics[width=\textwidth]{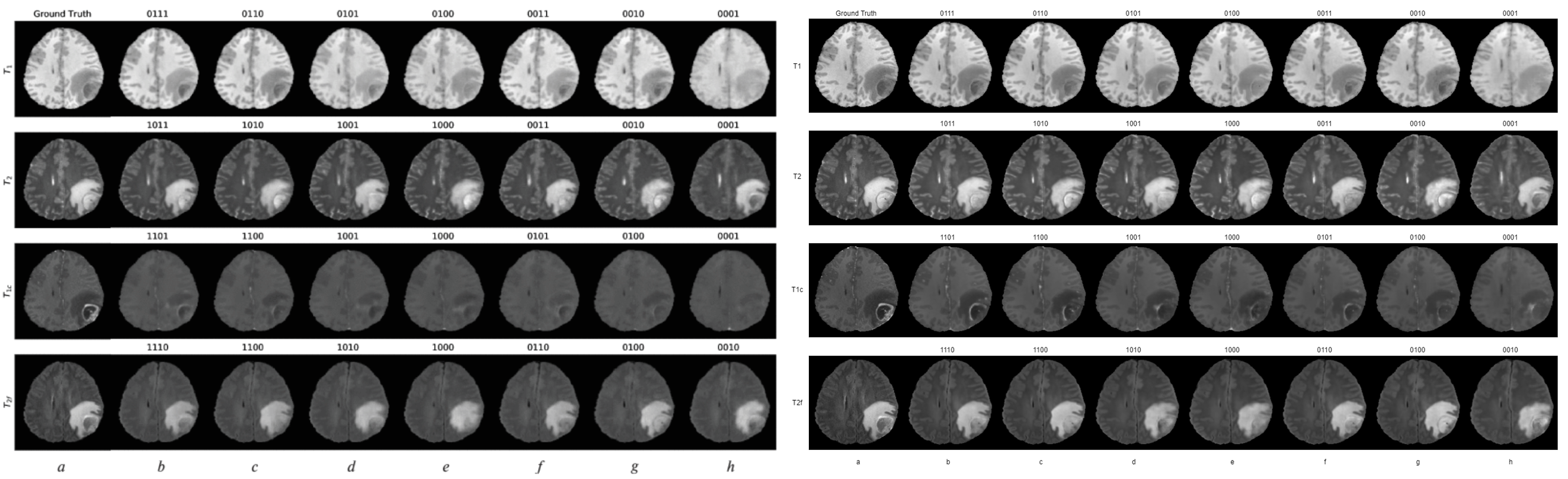}
\caption{This is an example (Brats18\_CBICA\_AAP\_1) of the synthesis results using the original and our implementation of mmGAN, which are quite visually similar to each other. Each row corresponds to a particular sequence (row names on the left in order T1, T2, T1c, and T2flair). Columns are indexed at the bottom of the figure by alphabets (a) through (h), and have a column name written on top of each slice. Column names are 4-bit strings where a zero (0) represents missing sequence that was synthesized, and one (1) represents presence of sequence. Column (a) of each row shows the ground truth slice, and the subsequent columns ((b) through (h)) show synthesized versions of that slice in different scenarios. The order of scenario bit-string is T1, T2, T1c, T2flair. For instance, the string 0011 indicates that sequences T1 and T2 were synthesized from T1c and T2flair sequences.} 
\label{fig:Brats18CBICAAAP1}
\end{figure}

This subsection compares the original implementation of mmGAN \textsuperscript{\ref{ft2}} 
and ours. In particular, we need to verify if our mmGAN can reproduce the consistent results as the original mmGAN, before we integrate with DeepMedic for brain tumor segmentation.
\noindent
We have presented our validation result using BRATS18 HGG dataset in Table \ref{ori_rst}. Our result is generally very close to the original one. This can be further visualised in Figure \ref{fig:mse_ssim_psnr_diff} using the exact difference values. Although, the largest difference is shown when only $T2_{flair}$ is missing, which corresponds to the scenario ``1110'', the absolute difference in terms of mean square error (MSE), peak signal to noise ratio (PSNR), and structured similarity indexing method (SSIM) are still marginal. Moreover, in Figure \ref{fig:Brats18CBICAAAP1} we have also presented comparative visualisation results to illustrate the small difference between MRI sequences generated by the original implementation and ours using a specific case (patient number: Brats18\_CBICA\_AAP\_1). Therefore, our mmGAN can reproduce the results by the original one according to the public available resources\cite{sharma2019missing}, \cite{menze2014multimodal}.

\begin{table}
\centering
\caption{Comparative results over HGG dataset of BRATS2018 for reproducing mmGAN (The order of sequences in scenario is T1, T2, T1c, T2f. For example, 0001 means only T2f is valid, other sequences are missing.).}\label{ori_rst}
\begin{tabular}{|l|ll|ll|ll|}
\hline
scenario & MSE-org & MSE-ours & PSNR-org & PSNR-ours & SSIM-org & SSIM-ours \\
\hline
0001 & 0.0143   & 0.0107  & 23.196   & 23.4940   & 0.8973    & 0.9007   \\
0010 & 0.0072   & 0.0086  & 24.524   & 24.1919   & 0.8984    & 0.9052   \\
0100 & 0.0102   & 0.0121  & 23.469   & 22.9292   & 0.9074    & 0.9033   \\
1000 & 0.0072   & 0.0097  & 24.879   & 23.6690   & 0.9091    & 0.9018   \\
0011 & 0.0060   & 0.0055  & 25.863   & 26.1124   & 0.9166    & 0.9332   \\
0101 & 0.0136   & 0.0108  & 22.900   & 23.9051   & 0.9156    & 0.9211   \\
0110 & 0.0073   & 0.0087  & 24.792   & 24.4054   & 0.9140    & 0.9182   \\
1001 & 0.0073   & 0.0069  & 26.189   & 25.3669   & 0.9264    & 0.9259   \\
1010 & 0.0040   & 0.0075  & 26.150   & 24.4325   & 0.9107    & 0.9069   \\
1100 & 0.0068   & 0.0091  & 25.242   & 24.0843   & 0.9175    & 0.9103   \\
0111 & 0.0091   & 0.0072  & 24.173   & 25.9732   & 0.9228    & 0.9436   \\
1011 & 0.0017   & 0.0031  & 28.678   & 27.2154   & 0.9349    & 0.9404   \\
1101 & 0.0098   & 0.0090  & 24.372   & 24.8936   & 0.9239    & 0.9241   \\
1110 & 0.0033   & 0.0084  & 26.397   & 23.6391   & 0.9150    & 0.9016   \\
mean & 0.0082   & 0.0084  & 24.789   & 24.5937   & 0.9120    & 0.9169   \\
\hline
\end{tabular}
\end{table}

\subsection{ACN and mmDM (mmGAN+DeepMedic)}
\begin{figure}[htb]
\centering
\includegraphics[width=\textwidth]{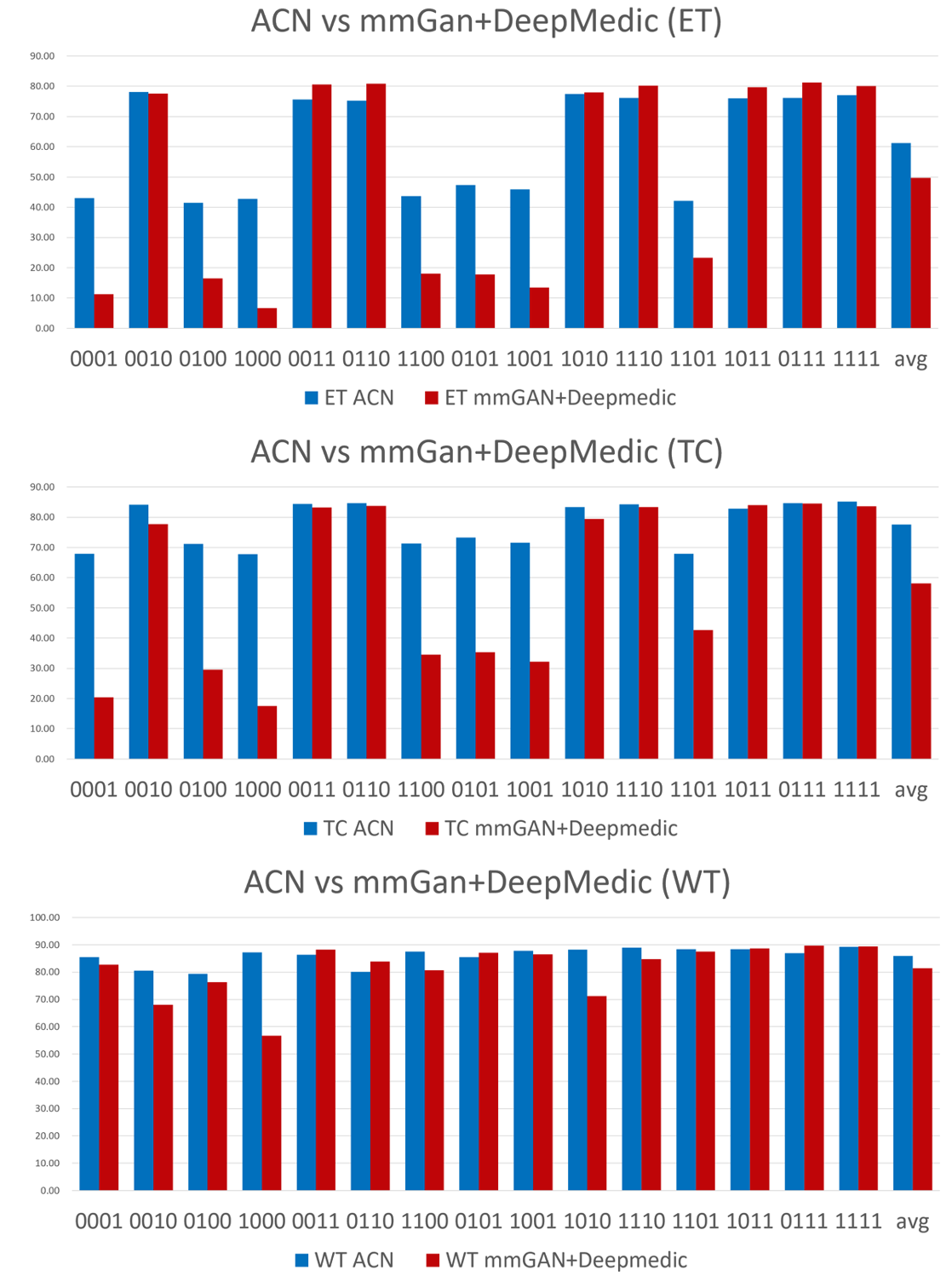}
\caption{ACN vs mmGan+DeepMedic (ET, TC, WT). This is the visualization version of evaluation results shown in Table \ref{acn_table}} 
\label{fig:acn_mmdm}
\end{figure}

Since mmGAN can synthesize missing MRI sequences with decent quality, can this really improve the performance of the downstream workflow such as brain tumor segmentation for diagnosis? State-of-the-art solutions such as ACN\cite{wang2021acn} can deal with brain tumor segmentation with missing MRI modalities without re-producing these missing MRI information as intermediate results. But can such a specialised model ACN always generate overwhelmingly better results than the straightforward combination (we call it mmDM for short) of mmGAN and DeepMedic \cite{kamnitsas2016deepmedic} (a state-of-the-art model for brain tumor segmentation)? This subsection attempts to answer these questions or at least initiates discussions with some preliminary results.
\noindent
As shown in Table \ref{acn_table} and Figure \ref{fig:acn_mmdm}, ACN is much better than mmDM especially when T1c MRI sequence is missing. However, in many cases, mmDM can provide comparable performance and sometimes even better than ACN (e.g., 1011). Moreover, we have found that for the enhancing tumor segmentation (ET), there are seven out of total fourteen possible missing modalities cases where the dice score is even lower than 50. These results are not high enough to confidently convince the doctor to use in clinics treatment or diagnosis.

\begin{table}
\centering
\caption{Comparing ACN and mmDM(mmGAN+DeepMedic) in dice score using BraST 2018 dataset HGG. (The order of modalities: T1, T2, T1c, T2f. For example, 0001 means only T2f is valid, other sequences are missing. ET: The Enhancing Tumor; TC: The Tumor Core; WT: The Whole Tumor). The visualization of this table results is shown in Figure \ref{fig:acn_mmdm}}\label{acn_table}
\begin{tabular}{cclclcl}
Type       & \multicolumn{2}{c}{ET}  & \multicolumn{2}{c}{TC}  & \multicolumn{2}{c}{WT}  \\
\hline
Modalities & ACN   & mmDM & ACN   & mmDM & ACN   & mmDM \\
\hline
0001       & 42.98 & 11.27           & 67.94 & 20.36           & 85.55 & 82.78           \\
0010       & 78.07 & 77.55           & 84.18 & 77.75           & 80.52 & 68.07           \\
0100       & 41.52 & 16.50           & 71.18 & 29.53           & 79.34 & 76.31           \\
1000       & 42.77 & 6.66            & 67.72 & 17.52           & 87.30 & 56.66           \\
0011       & 75.65 & 80.61           & 84.41 & 83.21           & 86.41 & 88.25           \\
0110       & 75.21 & 80.82           & 84.59 & 83.69           & 80.05 & 83.94           \\
1100       & 43.71 & 18.00           & 71.30 & 34.50           & 87.49 & 80.71           \\
0101       & 47.39 & 17.79           & 73.28 & 35.27           & 85.50 & 87.05           \\
1001       & 45.96 & 13.45           & 71.61 & 32.20           & 87.75 & 86.53           \\
1010       & 77.46 & 77.97           & 83.35 & 79.44           & 88.28 & 71.23           \\
1110       & 76.16 & 80.19           & 84.25 & 83.34           & 88.96 & 84.75           \\
1101       & 42.09 & 23.30           & 67.86 & 42.62           & 88.35 & 87.53           \\
1011       & 75.97 & 79.71           & 82.85 & 84.01           & 88.34 & 88.64           \\
0111       & 76.10 & 81.21           & 84.67 & 84.49           & 86.90 & 89.64           \\
1111       & 77.06 & 80.12           & 85.18 & 83.62           & 89.22 & 89.41           \\
avg        & 61.21 & 49.68           & 77.62 & 58.10           & 85.92 & 81.43     
\end{tabular}
\end{table}

\section{Conclusion and Future Work}
This technical report presents an improved Tensorflow implementation of mmGAN in terms of efficiency and usability. Moreover, this technical report analyses preliminary results for brain tumor segmentation with missing MRI sequences. We conclude that state-of-the-art solutions can not guarantee good segmentation results in many missing MRI sequences cases. Sometimes it is close to or even worse than a simple combined version of mmGAN and DeepMedic. Future work should focus on defining a theoretic upper bound on how well the model can perform given restricted information. This is because too much information lost can not lead to a great result in the end. Additionally, closer interdisciplinary cooperation between computer scientists and doctors should be highly promoted to solve the challenge in brain tumor segmentation under missing MRI sequences.




\bibliographystyle{splncs04}
\bibliography{tech_rep}

\end{document}